\documentclass[letterpaper, 10 pt, conference]{ieeeconf}  

\IEEEoverridecommandlockouts                              

\usepackage{graphics} 
\usepackage[pdftex]{graphicx}
\usepackage{amsmath} 
\usepackage{cite}
\usepackage{subfig}
\usepackage{float}
\usepackage{multirow}
\usepackage{tabularx}
\usepackage{mathtools}
\usepackage{amssymb}
\usepackage{balance}

\title{\LARGE \bf
Radar-on-Lidar: metric radar localization on prior lidar maps
}

\author{Huan Yin$^{1}$, Yue Wang$^{1}$, Li Tang$^{1}$ and Rong Xiong$^{1}$
\thanks{$^{1}$ All authors are with the State Key Laboratory of Industrial Control and Technology, and the Institute of Cyber-Systems and Control, Zhejiang University, Hangzhou 310058, China.}%
\thanks{This work was supported in part by the National Key R\&D Program of China (2018YFB1309300), and in part by the National Nature Science Foundation of China (61903332), and in part by the Key R\&D Program of Zhejiang Province (2019C01043).}
}

\begin{document}

\maketitle
\thispagestyle{empty}
\pagestyle{empty}

\begin{abstract}
Radar and lidar, provided by two different range sensors, each has pros and cons of various perception tasks on mobile robots or autonomous driving. In this paper, a Monte Carlo system is used to localize the robot with a rotating radar sensor on 2D lidar maps. We first train a conditional generative adversarial network to transfer raw radar data to lidar data, and achieve reliable radar points from generator. Then an efficient radar odometry is included in the Monte Carlo system. Combining the initial guess from odometry, a measurement model is proposed to match the radar data and prior lidar maps for final 2D positioning. We demonstrate the effectiveness of the proposed localization framework on the public multi-session dataset. The experimental results show that our system can achieve high accuracy for long-term localization in outdoor scenes. 

\end{abstract}

\section{Introduction}

Localization on map is an essential and critical component for autonomous navigation systems, which estimates the precise metric position for mobile robots or vehicles as an assistance. In recent years, localization algorithms and methods are relatively mature and success with the development of sensor technology and back-end optimization. However, the reliable localization is still challenging for long autonomy, due to the unpredictable conditions on roads, including the variable of illumination and weather etc.  

To overcome these constraints of variations, different types of sensors are selected or connected on board, according to the requirements of robots and relative operating environments. For instance, pure vision systems were proposed and popular for indoor localization \cite{mur2017orb}. As for field robots, visual descriptors and features change a lot from various illumination, thus lidar based methods were widely used for metric positioning across day and night \cite{krusi2015lighting}, including laser map aided visual localization \cite{ding2019persistent}. The success of lidar based mapping and localization enabled the application of mobile robots in real world. But the weather invariance still remains challenging, especially in foggy or snowy days, some scenerios may not return any meaningful measurements for laser scanner, leading to a potential difficulty for long-term localization.

Radio detection and ranging (Radar) is considered as a weather-invariant on board sensor, which provides distance and speed measurements of vehicles for driver assistance systems. Recent scanning radar sensors achieve much improvements with higher resolution and broader field of view \cite{RadarRobotCarDatasetArXiv, yspark-2019-icra-ws}, but various types of noises still remains a significant challenge for applications, including ghost detections, multi-path reflection etc \cite{vivet2013localization, lu2019millimap}. Many researches tried to filter the noisy radar data to a more precise form by geometrical or learning methods, and majority of the works mainly focus on the radar data processing.

\begin{figure}[t]
	\centering
	\includegraphics[width=8cm]{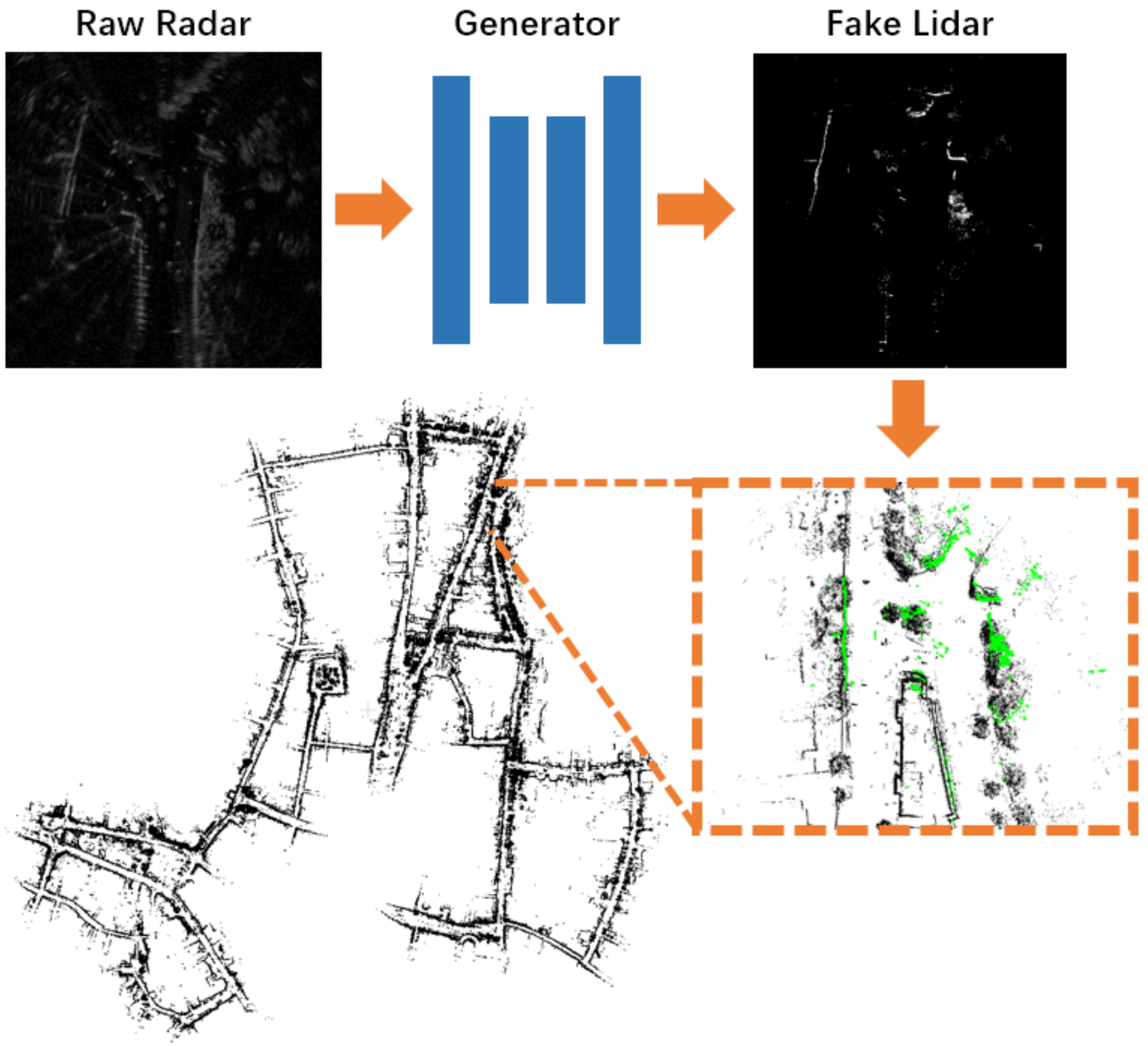}
	\caption{ A brief illustration of the proposed localization system. The raw radar is first transfered to lidar-like representation by the trained generator in GAN. Then the fake lidar points are transformed into MCL system for pose estimation. The prior 2D lidar map for localization is shown in black, and the green points stand for the transformed fake lidar on the estimated pose. }
	\label{introduction}	
\end{figure} 

In this paper, we consider that lidar sensor can be a connection for radar application in real world, since both of the sensors provide range sensing on mobile platforms. Specifically, range data from laser scanner are more accurate and reliable compared to radar, so we use a conditional generative adversarial network (GAN): pix2pix \cite{isola2017image}, which can learn the lidar representations for Frequency-Modulated Continuous-Wave (FMCW) radar data. The trained generative model is able to transfer the radar data to fake lidar points. Then a Monte Carlo localization (MCL) system is formulated by motion and measurement models. The whole system can localize the robot with the transfered data on a pre-built laser map, illustrated in Fig.~\ref{introduction} and Fig.~\ref{system}. Overall, the contributions of this paper are as follows:
\begin{itemize}
\item Conditional GAN is used to transfer the radar data to fake lidar points, which has a more precise form compared to raw data.
\item We propose a Monte Carlo based localization system to localize the mobile robot, with radar data input and pre-built lidar maps.
\item Multi-session dataset is employed to demonstrate the effectiveness of the proposed system across days.
\end{itemize} 

The rest of this paper is organized as follows: In
Section \uppercase\expandafter{\romannumeral2}, we introduce the related work about radar based localization. The proposed localization system is presented in Section \uppercase\expandafter{\romannumeral3}. Section \uppercase\expandafter{\romannumeral4} reports the experimental results using public dataset. And a brief conclusion and future outlook are addressed in Section \uppercase\expandafter{\romannumeral5}. 

\section{Related Work}

Multi-sensor based localization is widely applied in the area of mobile robot or intelligent vehicles. We mainly focus on the radar based localization in this section to review the recent development. 

Radar data processing is a key part for real application in real world since there are many noises and Doppler effect in the raw radar data. Some researchers studied the Doppler effect and data distortion formulation in order to estimate the robot's displacement \cite{vivet2013localization}. Radar data clustering was used in \cite{schuster2016robust}, which can achieve high accuracy with wheel odometry from vehicle. Similarly, feature extraction method was used in \cite{schuster2016landmark}, which detected distinguishable landmarks for radar based simultaneous localization and mapping (SLAM).

With the development of FMCW radar sensor, two new datasets are presented for researching: Oxford Radar RobotCar Dataset\cite{RadarRobotCarDatasetArXiv} and MulRan Dataset for Urban Place Recognition \cite{yspark-2019-icra-ws}. The Navtech Millimetre-Wave FMCW radar is able to achieve more accurate range sensing, and several radar based research works for outdoor scenes are also conducted on these two datasets. One of the topics is the ego-motion estimation between two radar scans, which is published in many recent papers \cite{cen2018precise, aldera2019fast, cen2019radar, barnes2019masking}. These methods proposed effective algorithms to filter the noisy radar data and estimate the transformation between two radar scans. Conventional signal processing technique was used to detect the landmarks from radar data \cite{cen2018precise}, and then data association was performed for scan matching. Keypoints based method was also proposed to achieve real-time radar odometry \cite{cen2019radar}. In addition, data filter can be achieved by learning based methods. In \cite{aldera2019fast}, ground truth poses were used to annotate the raw data pixel by pixel by accumulating radar scans, and then U-Net was trained under the supervision of the labels. The trained U-Net is able to discard the useless radar points and keep the high power points for motion estimation. End-to-end radar odometry \cite{barnes2019masking} was proposed to replace the conventional scan matching methods, which can also perform robust and efficient odometry system. But due to the limitation of motion estimation, some errors may occur during the robot moving. In \cite{aldera2019could}, a failure detection method was proposed to improve the reliability of odometry system. 

The connection between lidar and radar is a popular topic for range sensing in recent years. Place recognition is a critical module in localization and mapping system. Compared to lidar based place recognition or loop closing, there are few research search on radar range data. In \cite{gskim-2020-mulran}, researchers find that ScanContext \cite{kim2018scan} for lidar place recognition can also be used for radar data. A geometrical solution for indoor localization was shown in \cite{park2019radar}, which registered the radar points on CAD models and lidar maps. The relationship between lidar and radar was also performed in \cite{weston2019probably}. An Inverse Sensor Model was learned from lidar data by neural network, which can filter the noises and occlusions in radar data. Considering that lidar and radar are both range sensors, we use the neural network to transfer the radar data to lidar points directly in this paper. With the output from the trained network, a localization system on lidar maps is built for pose estimation.

\section{Methods}

\begin{figure*}[t]
	\centering
	\includegraphics[width=18cm]{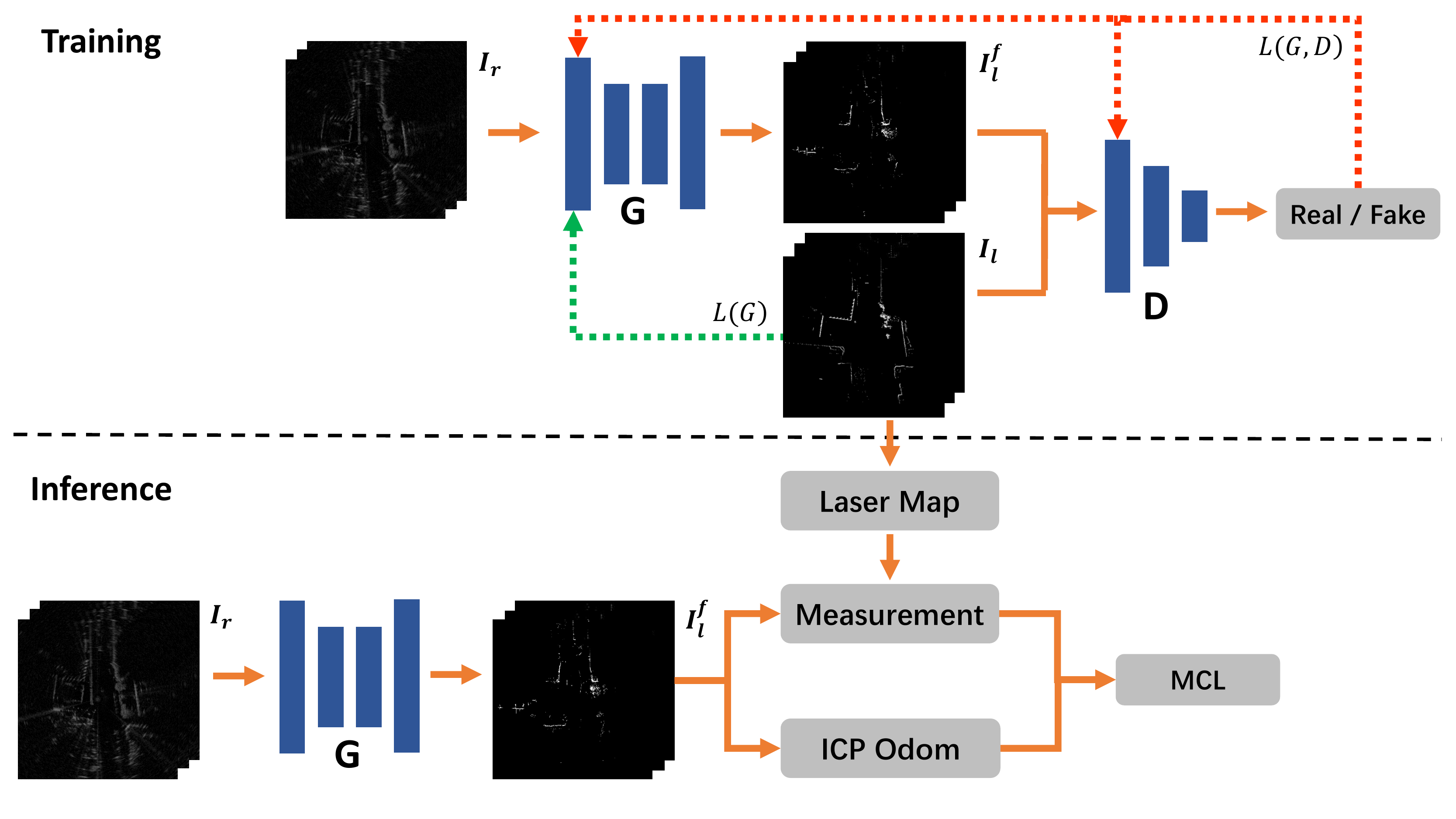}
	\caption{ The framework of the proposed localization system. In the training session, we use a GAN network to achieve radar data transfer, which contains two objectives: $\mathcal{L}(G,D)$ and $\mathcal{L}(G)$, shown in the figure. In the inference session, the trained generator is used to transform the radar data to fake lidar data for localization. The components are components of MCL are ICP registration for odometry estimation, and the a measurement model on laser map. }
	\label{system}	
\end{figure*}

The proposed system is presented in this section, including the pix2pix network for data filter, and the Monte Carlo localization system. The whole system is shown as Fig.~\ref{system}. 

\subsection{Radar-Lidar style transfer}

In this paper, the FMCW radar data or lidar data in 2D-XY cartesian coordinate has two representations for processing: the 2D image-like representations $I$ for neural network training, and 2D point cloud $P$ for metric localization. Specifically, images ${I}$ are used for image style transfer by deep learning, and point clouds ${P}$ are used for metric localization of mobile robot. Both of these two representations can be transformed to each other under a certain resolution $r$, as follows:
\begin{equation} \label{image_pc}
I \rightleftharpoons P
\end{equation}
\par Given a sequence of radar and lidar images from bird-eye view, donated as $I_{r}$ and $I_{l}$, we first align the lidar data on the radar coordinate by calibration results, and find the nearest lidar scan of each radar by timestamps. With the aligned pairwise data, we train a pix2pix network \cite{isola2017image} to learn the lidar representation for radar data. The objective of the pix2pix network contains two parts, as follows:
\begin{equation} \label{loss}
G = {\rm arg \mathop{min}\limits_{G} \mathop{min}\limits_{D}} \mathcal{L}(G,D) + \lambda\mathcal{L}(G)
\end{equation}
which are the objective of conditional GAN and the L1 distance loss. The loss of generator and discriminator is formulated as follows:
\begin{equation} \label{loss_1}
\begin{aligned} 
\mathcal{L}(G,D) = &E_{(I_{r}, I_{l})}[{\rm log} D(I_{r}, I_{l})]+ \\
&E_{(I_{r})}[{\rm log}(1-D(I_{r},G(I_{r})))]
\end{aligned}
\end{equation}
and the $L_1$ distance of ``real'' lidar image and ``fake'' image is as follows:
\begin{equation} \label{loss_2}
\mathcal{L}(G) = E_{(I_{r}, I_{l})}[\left \| I_{l} - G(I_{r}) \right \|_{1}]
\end{equation} 
\par Overall, as shown in Fig.~\ref{system}, pix2pix network is able to train a generator $G$ and discriminator $D$ after learning lidar style transfer, and only $G$ is used in the inference step,  the following metric localization on lidar maps. 

Generally, the ground in $x-y$ planes in 3D point clouds are used to constrain the Z-axis for 6DoF localization. While in this paper, we consider the ground planes are redundant for 2D positioning, and we use a height threshold on original 3D lidar point cloud for filtering ground planes, thus a more precise 2D lidar representation can be used for radar style transfer and following operations. Finally, the filtered 2D lidar data are fed into GAN for training, then we can obtain the fake lidar image and point cloud by using the trained generator, as follows:
\begin{equation} \label{transfer}
G(I_{r}) \mapsto I_{l}^{f}, P_{l}^{f}
\end{equation}

\subsection{Monte Carlo localization}

With the generated fake lidar point clouds, the metric localization can be achieved by registration on the laser map directly. Considering the uncertainty of neural networks, there still exist some noises on the fake lidar compared to the real lidar data, so a reliable back-end is required for long-term localization. In this paper, Monte Carlo localization, or particle filter localization, is used as the pose estimation method. MCL is widely used mobile platforms due to its pose tracking and re-localization abilities \cite{thrun2002probabilistic, yin20193d}.

In this paper, we follow the two steps of conventional MCL method: the prediction phase by motion model and the update phase by measurement model. Based on a prior laser map $M$,  the particle filter is the following factorization:
\begin{equation} \label{MCL}
\hat{p}(x_t | Z_t, M) = \eta p(z_t | x_t, M) \hat{p}(x_t | Z_{t-1}, M) 
\end{equation}
where $\eta$ is a normalizer. And the measurements $Z_t=\left\lbrace z_1,...,z_t \right\rbrace =\left\lbrace P_1,...,P_t\right\rbrace $, in which the fake lidar cloud $P_l^f$ is denoted as $P$ for clearance.

In the prediction phase, the robot state is approximated by samples as follows:
\begin{equation} \label{prediction}
\hat{p}(x_t | Z_{t-1}, M) = \sum_i p(x_t | s_{t-1}^i, u_{t-1}) 
\end{equation}
where $s^i$ represents a particle in estimator, and $u$ is the robot motion, which is achieved by applying 2D iterative closest points (ICP) method to align $P_{t-1}$ and $P_{t}$. 

In the update phase, to calculate the likelihhood of $s_t^i$ given $z_t$ on $M$, the individual importance weight $\omega_t^i = p(z_t | s_t^i, M)$ is needed. We define the weight as a power function relationship to the number of matched points $N_t^i$ in point cloud $P_t^i$, as follows:
\begin{equation} \label{weight}
\omega_t^i = (N_t^i)^\lambda 
\end{equation}
where $\lambda$ is a constant value, and $P_t^i$ is the point cloud $P_t$ aligned on $s_t^i$. A direct form to determine whether a radar point is ``matched'' can be formulated by the nearest distance to the map points: if a point is close the map point, we consider it contributes to the localization effectively. Based on a threshold distance $d_{th}$ and the nearest distance $d_{p,M}$, we set a radar point as matched with the following criteria:
\begin{equation}\label{observation}
matched = \left\{\begin{array}{cc}
0 & d_{p, M} \geq d_{th}  \\
1 &d_{p, M} < d_{th}
\end{array}, p \in P_t^i
\right.
\end{equation}
which is similar with the observation count for map points in \cite{yin20193dl}. Based on the distance statistics on the $P_t^i$, the importance weights can be obtained. The particles with larger $N_t^i$ and higher likelihood are kept after weight normalization and re-sampling step, and the robot pose is estimated combing with the particles' states and weights. Finally, the prediction and update phases are repeated recursively with the input lidar scans from the trained generator model. 


\begin{figure*}[t]
	\centering
	\includegraphics[width=18cm]{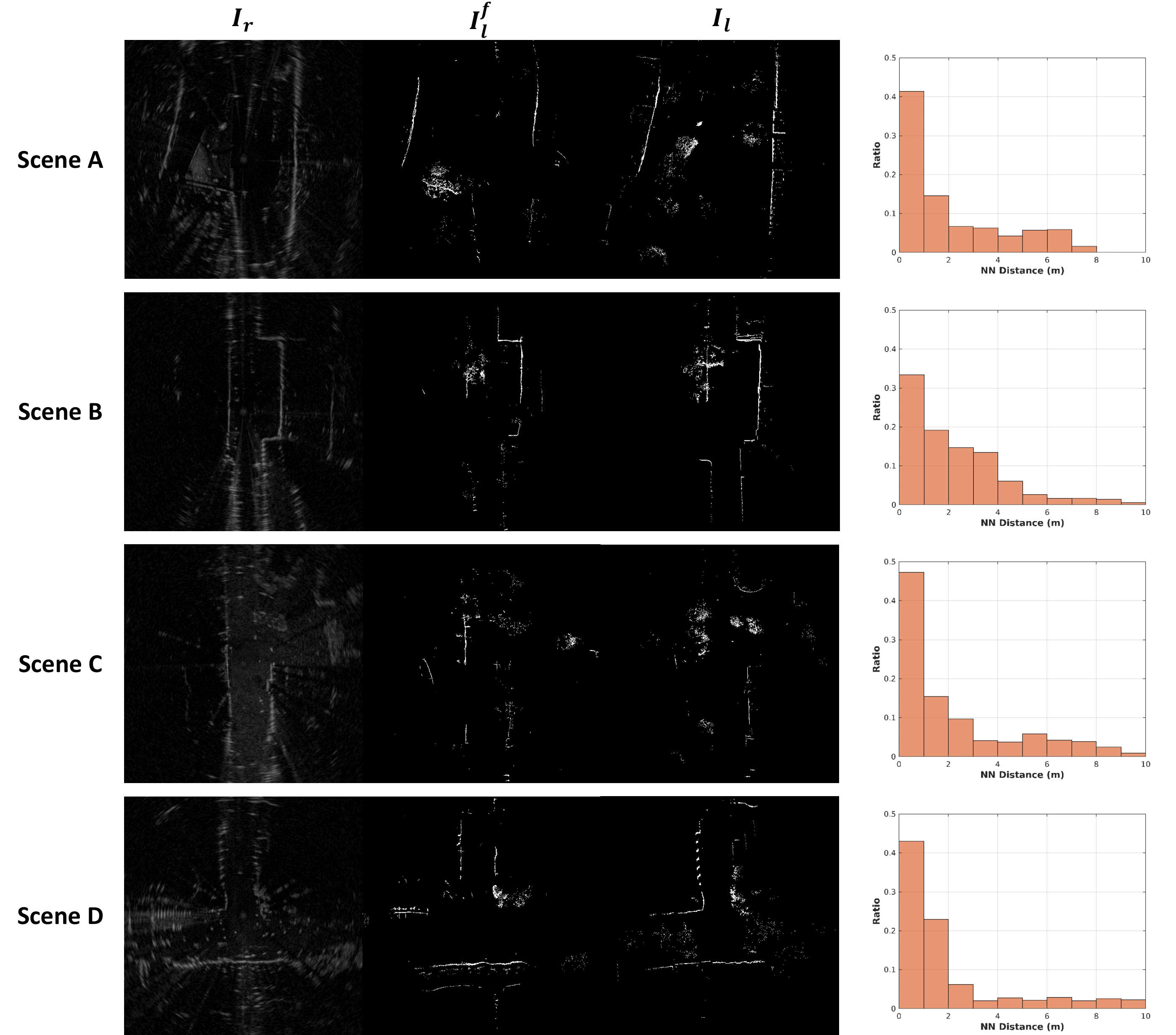}
	\caption{ We present four cases of image style transformation and similarity measurement as above. The raw radar data $I_r$, the generated fake lidar $I_l^f$, the real lidar $I_l$ and the similarity measurement are shown from left to right in each row. The histogram shows the distribution of nearest distances from $P_l^f$ to $P_l$. }
	\label{images}	
\end{figure*} 

\section{Experiments}

Experimental results are presented in this section with public dataset. We first introduce the dataset and some preparations, and then present the generated fake lidar point clouds. Finally, localization performance is evaluated on multiple sessions. To help understanding our experimental results, a supplimentary video is uploaded \footnote{https://youtu.be/wE-DbaCDbsU}.

\subsection{Dataset and implementation}

We use the public multi-session dataset, Oxford Radar RobotCar Dataset \footnote{https://dbarnes.github.io/radar-robotcar-dataset} \cite{maddern20171, RadarRobotCarDatasetArXiv} for performance evaluation, which includes dual Velodyne HDL-32E lidar sensors and one Navtech CTS350-X Millimetre-Wave FMCW radar. The calibration results and optimised ground truth radar odometry are also provided. 

The provided ground truth odometry can be used for ego-motion or odometric error evaluation. But for the localization across days, the starting position is different for each single session, so the ground truth odometry can not be used directly in this paper. To achieve proper evaluation, we first select five sessions with the same starting place, and set the initial state as zero for each one. Then five laser maps are accumulated by using the dual lidar data and ground truth odometry with initial states. Finally, we set the first session map as the target global map, and use a precise ICP method to register other maps on it. The ground truth poses of each session can be obtained by multiplying each global ICP result, which are appropriate for localization evaluation under the same global coordinate. And we present the detailed information of these selected sessions in Oxford Radar RobotCar Dataset, shown in Tab.~\ref{dataset}.

As for the image translation, we use the open code of pix2pix network \footnote{https://github.com/phillipi/pix2pix}. The generator network is set as ResNet with 9 blocks and default configurations. We set the one-channel range images with $512\times512$ size and resolution of $0.25m / pixel$. The images are handled with original sizes and no crop or resize operations. 

\subsection{Real-Fake image comparison}

For localization application, we first transfer the radar image $I_{r}$ to lidar image $I_{l}$, which filter the noisy radar data to fake lidar data. Some scenes with images are presented in Fig.~\ref{images}. Intuitively, the generated images are similar with the lidar image, but there are still some noises after processing; and fewer lidar points are remained through network compared to the original lidar data.   

To compare the two images $I_l^f$ and $I_l$ quantitatively, we use a statistic based method to measure the similarity between the point clouds $P_l^f$ and $P_l$. We build a K-Dimensional tree on the real lidar point clouds, and then every point in $P_l^f$ obtains its nearest neighbor (NN) distance. All the NN distances are counted as histograms for similarity measurement, as shown in Fig.~\ref{images}. Almost more than 40\% points in $P_l^f$ are close to the lidar points and the distances are below $1m$, and nearly 60\% points when the distance is $2m$. Specifically, the generated points are mainly focused on the bushes, walls and trees on the sides of the road. These lidar points are informative for metric localization, but there are still some ghost objects in the fake lidar.

\begin{table}[t]
	\captionsetup{justification=centering}
	\renewcommand\arraystretch{1.5}
	\begin{center}
		\caption{ Sessions for evaluation}
		\label{dataset}
		\begin{tabular}{p{1.2cm}p{0.8cm}p{1.0cm}p{1.0cm}p{2.6cm}}
			\hline
			Date & Times& Duration & Distance & Usage \\
			\hline
			10/01/2019& 11:46:21& 00:37:00& 9.02 $km$& Training and Mapping \\
			\hline
			10/01/2019& 12:32:52& 00:35:57& 9.04 $km$& Seq01 for localization \\
			11/01/2019& 12:26:55& 00:32:00& 9.03 $km$& Seq02 for localization \\
			16/01/2019& 13:42:28& 00:31:53& 9.01 $km$& Seq03 for localization \\
			18/01/2019& 12:42:34& 00:34:13& 9.03 $km$& Seq04 for localization \\
			\hline
		\end{tabular}
	\end{center}
\end{table}

\begin{table}[!t]
	\captionsetup{justification=centering}
	\renewcommand\arraystretch{1.5}
	\begin{center}
		\caption{ Localization results: RMSE error}
		\label{localization_1}
		\begin{tabular}{p{1.2cm}p{2.0cm}p{2.0cm}p{2.0cm}}
			\hline
			Sequence & Failed Times& Positional RMSE& Yaw RMSE \\
			\hline
			seq01& 0& 8.46$m$& 5.43$^\circ$ \\
			seq02& 0& 6.93$m$& 2.46$^\circ$ \\
			seq03& 0& 9.12$m$& 4.47$^\circ$\\
			seq04& 2& 14.10$m$& 4.25$^\circ$\\
			\hline
		\end{tabular}
	\end{center}
\end{table}

\begin{table}[!t]
	\captionsetup{justification=centering}
	\renewcommand\arraystretch{1.5}
	\begin{center}
		\caption{ Localization results: error distribution}
		\label{localization_2}
		\begin{tabular}{p{1.2cm}p{2.0cm}p{2.0cm}p{2.0cm}}
			\hline
			Sequence & \textless $1m$, $2^\circ$& \textless $2m$, $5^\circ$& \textless $5m$, $10^\circ$ \\
			\hline
			seq01& 29.55\%& 59.89\% & 74.74\% \\
			seq02& 27.22\%& 68.08\% & 86.25\% \\
			seq03& 13.71\%& 48.47\% & 77.10\% \\
			seq04& 23.57\%& 48.49\% & 68.56\% \\
			\hline
		\end{tabular}
	\end{center}
\end{table}

\subsection{Localization results}

We conduct the localization test on four sessions from Oxford Radar RobotCar Dataset, shown in Tab~\ref{dataset}. These sessions traveled a distance over $35km$ across eight days in urban area, which can validate the effectiveness of the proposed localization system under challenging environments. For Seq01, Seq02 and Seq03, the proposed system can achieve complete results without localization failure when running. While for Seq04, the estimator fails at two places when localizing; so we re-set the initial state at these two places manually, and then continue the localization process.

\begin{figure*}[t]
	\centering
	\includegraphics[width=18cm]{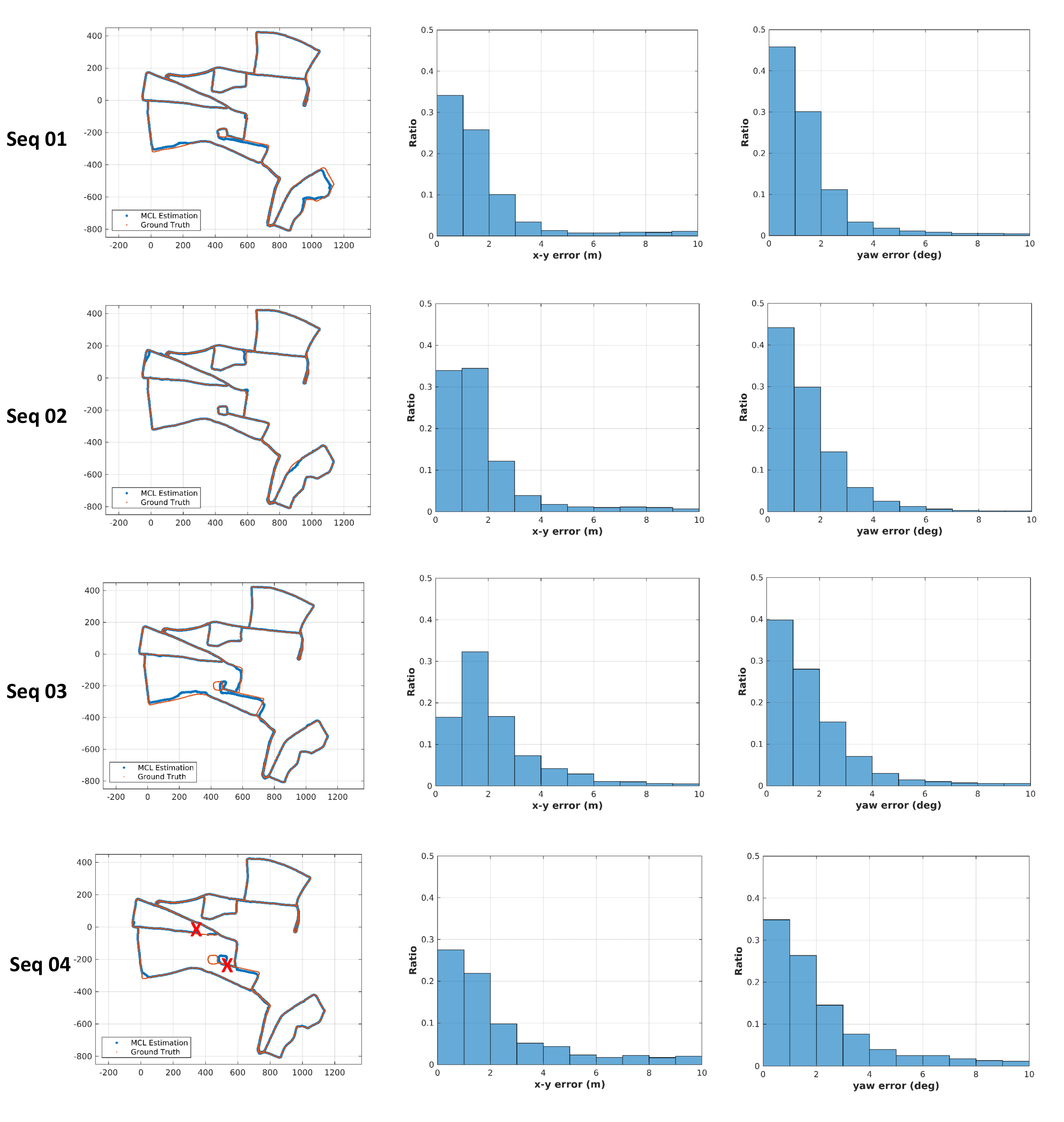}
	\caption{ The localization evaluation in different days. The trajectories, and the histograms of position and heading are shown from left to right in each row.}
	\label{localization_3}	
\end{figure*}

To present the results precisely, we calculate the root mean square error (RMSE) of position and orientation, shown in Tab.~\ref{localization_1}; and the percentages of localization results within three thresholds are listed in Tab.~\ref{localization_2}. The trajectories and error distributions are displayed in Fig.~\ref{localization_3}. The majority of positional errors of all sessions are within $5m$, and the heading errors are limited in $3^\circ$. On the one hand, there are some large errors are at some cases, the vehicle running or turning at high speed for example, which have effects on the average error calculation. On the other hand, these large errors are reduced during traveling by re-localization, which validates the feasibility of the MCL system for long-term operation. 

Generally, we consider that the localization errors or failures are caused by the reasons as follows:
\begin{itemize}
\item The laser scanner has a smaller detecting range compared to the radar sensor, which makes the data annotation or image style transfer incomplete. Only the points near to the radar sensor ($\leq 64m$) can be used for localization in the experiments. 
\item The learned radar representations by neural networks are still noisy. The noises in fake lidar points mainly focus on ghost objects on roads, which are hard to distinguish from real landmarks in raw radar data.
\item The ICP registration in motion model is affected by some noisy data directly. As for measurement model, the pose estimation is determined by the number of particles and some other parameters, which makes the precise positioning difficult in challenging environments. 
\end{itemize}

\section{Conclusion}

This paper proposed a radar based localization system to estimate the robot poses on pre-built lidar maps. The whole system relies on the the image transfer network at the front-end and Monte Carlo localization at the back-end. Specifically, the trained GAN network is able to generate fake lidar point clouds from raw radar data; and then MCL is performed to localize the mobile robot. The effectiveness of the proposed system is validated on the Oxford Radar RobotCar Dataset. In the future, we consider that a more efficient and concise framework is desired by real-time computing for localization, which can be achieved by end-to-end deep learning technology in recent years.


\bibliographystyle{IEEEtran}
\balance
\bibliography{root}

\end{document}